\documentclass[sigconf]{acmart}

\AtBeginDocument{%
  }

\usepackage{bbm}
\usepackage{dsfont}
\usepackage{romannum}
\usepackage[unicode]{hyperref}
\usepackage{times}
\usepackage{epsfig}
\usepackage{graphicx}
\usepackage{amsmath}
\usepackage{hhline}
\usepackage{algorithm}
\usepackage{arydshln}
\usepackage{algpseudocode}
\usepackage{subcaption}
\usepackage{multirow}
\usepackage[flushleft]{threeparttable}
\usepackage{amsfonts}
\usepackage{booktabs}
\usepackage{siunitx}
\usepackage{makecell}
\usepackage{colortbl}

\usepackage{lipsum}

\newcommand\blfootnote[1]{%
  \begingroup
  \renewcommand\thefootnote{}\footnote{#1}%
  \addtocounter{footnote}{-1}%
  \endgroup
}

\newcommand{\reffig}[1]{Fig. \ref{#1}}
\newcommand{\refsec}[1]{Section \ref{#1}}
\newcommand{\reftab}[1]{Table \ref{#1}}

\begin{document}
\title{Frequency Regularization: Reducing Information Redundancy in Convolutional Neural Networks}

%\author{Paper ID: 1077}

\author{Chenqiu Zhao*}
\affiliation{%
	\institution{University of Alberta}
  \city{Edmonton}
  \country{Canada}}
\email{zhao.chenqiu@ualberta.ca}

\author{Guanfang Dong*}
\affiliation{%
	\institution{University of Alberta}
  \city{Edmonton}
  \country{Canada}}
\email{guanfang@ualberta.ca }

\author{Shupei Zhang*}
\affiliation{%
	\institution{University of Alberta}
  \city{Edmonton}
  \country{Canada}}
\email{shupei2@ualberta.ca}

\author{Zijie Tan}
\affiliation{%
	\institution{University of Alberta}
  \city{Edmonton}
  \country{Canada}}
\email{ztan4@ualberta.ca}

\author{Anup Basu}
\affiliation{%
	\institution{University of Alberta}
  \city{Edmonton}
  \country{Canada}}
\email{basu@ualberta.ca}
\renewcommand{\shortauthors}{Zhao et al.}

\begin{abstract}
	Convolutional neural networks have demonstrated impressive results in many computer vision tasks.
	However, the increasing size of these networks raises concerns about the information overload resulting from the large number of network parameters. 
	In this paper, we propose Frequency Regularization to restrict the non-zero elements of the network parameters in the frequency domain. 
	The proposed approach operates at the tensor level, and can be applied to almost all network architectures. 
	Specifically, the tensors of parameters are maintained in the frequency domain, where high frequency components can be eliminated by zigzag setting tensor elements to zero. 
	Then, the inverse discrete cosine transform (IDCT) is used to reconstruct the spatial tensors for matrix operations during network training. 
	Since high frequency components of images are known to be less critical, a large proportion of these parameters can be set to zero when networks are trained with the proposed frequency regularization. 
	Comprehensive evaluations on various state-of-the-art network architectures, including LeNet, Alexnet, VGG, Resnet, ViT, UNet, GAN, and VAE, demonstrate the effectiveness of the proposed frequency regularization.
	For a very small accuracy decrease (less than 2\%),
	a LeNet5 with 0.4M parameters can be represented by only 776 float16 numbers (over 1100$\times$ reduction), and a UNet with 34M parameters can be represented by only 759 float16 numbers (over 80000$\times$ reduction).
	In particular, the original size of the UNet model is 366MB, we reduce it to 4.5kb.
\end{abstract}


\maketitle

\section{Introduction}
\begin{figure}[h]
  \centering
  \includegraphics[width=\linewidth]{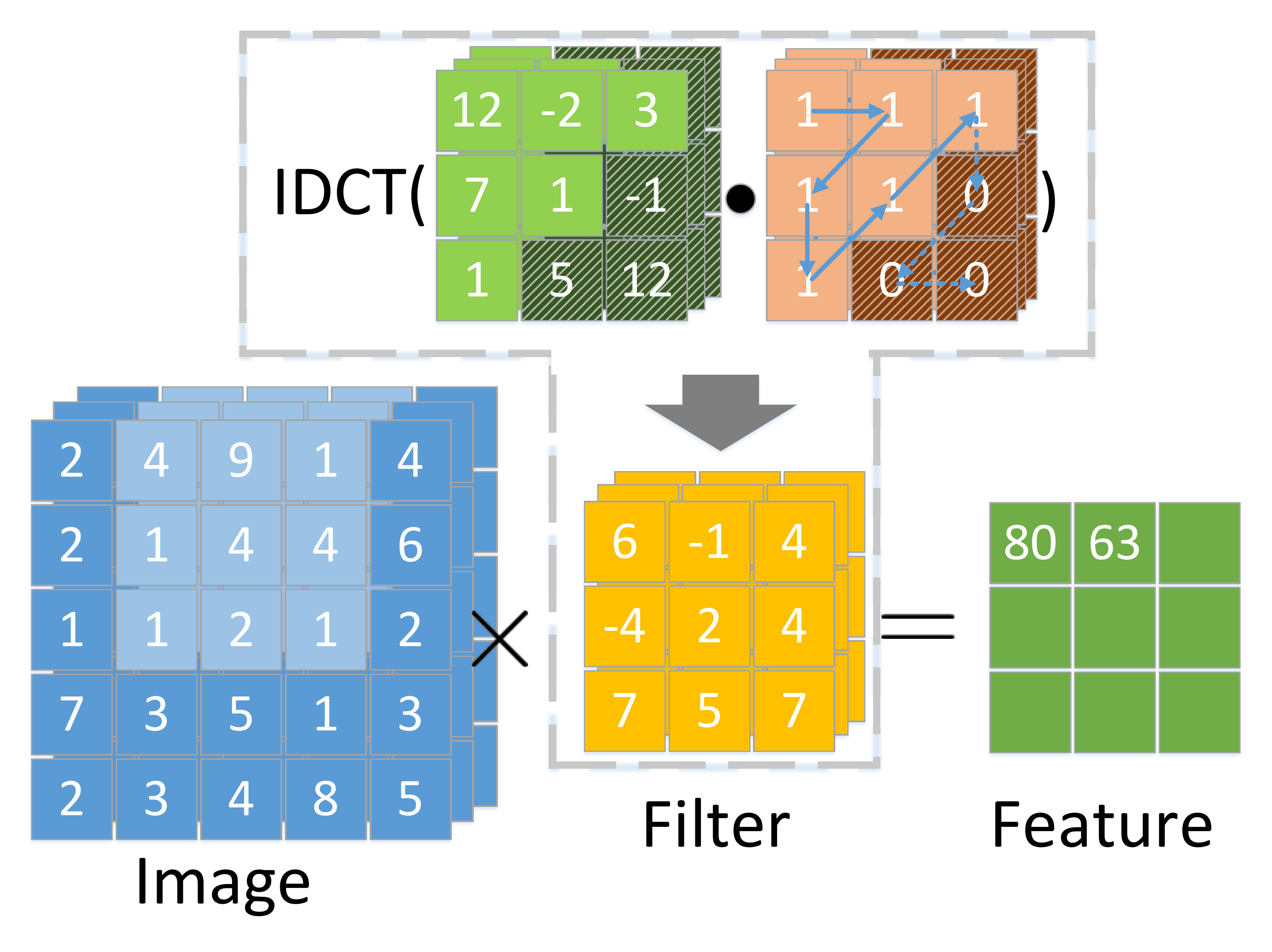}
	\vspace{-24pt}
	\caption{Illustration of the proposed frequency regularization. The tail elements of tensors in the frequency domain are truncated first, then input into the inverse of the discrete cosine transform to reconstruct the spatial tensor for learning features.}
	\label{fig_idea}
	\vspace{-20pt}
\end{figure}
\blfootnote{*Authors are equally contributed.}
Convolutional neural networks have become increasingly popular in computer vision applications, such as image classification, image segmentation and so on.
However, as the learning ability of these networks has improved, so has their size, growing from a few megabytes to hundreds of gigabytes \cite{dosovitskiy2021an}. 
This leads to challenges such as enormous storage space or long transmission time on the Internet.
After conducting a thorough literature review, 
we found that although larger networks tend to perform better, the accuracy improvement is not always directly proportional to the network size. 
In some cases, doubling the network size may not result in a significant accuracy improvement.
From this observation we conclude that there may be information redundancy within various network architectures, leading to the question: ``How can we reduce network information redundancy?''

It is commonly accepted that the performance of networks comes from the features learned by their parameters. 
For example, when a convolutional neural network such as Alexnet \cite{alexnet} is applied to an image classification task, 
the features learned by its convolutional kernels are closely related to the training images. 
Under this condition, features learned by a network are expected to have properties similar to the training images in which the high frequency components are known to be less important.
Thus, it is reasonable to apply the frequency domain transforms, such as the discrete cosine transform, to the network parameters. 
Unfortunately, given the poor interpretability of complex network architectures, even a small change in a few key parameters can significantly affect the network performance.
Based on this insight, the potential of frequency domain transforms for compressing or pruning networks may not be well-developed on pre-trained models, which has been the focus of previous approaches \cite{liuFrequencydomainDynamicPruning2018, 2020_cvpr_9157408, NIPS2016_36366388, 2019_TPAMI_8413170}. 
Instead, we focus on using the frequency domain transform as a network regularization to restrict information redundancy during the training process, and introduce Frequency Regularization (FR) as shown in \reffig{fig_idea}.

The proposed frequency regularization can be divided into two steps: dynamic tail-truncation and inverse discrete cosine transform (IDCT).
During network training, parameter tensors are maintained in the frequency domain, with the tail parts representing high frequency information zigzag truncated. 
The truncation process is implemented through a dot product with a zigzag mask matrix to ensure differentiability. 
After dynamic tail-truncation, tensors are input into the IDCT to reconstruct the spatial tensors that are then used as regular learning kernels in networks. 
Since the IDCT is a differentiable process, the actual tensors in the frequency domain can be updated through backpropagation algorithms. 
Moreover, as the reconstructed spatial tensors have the same size as those maintained in the frequency domain, 
the proposed frequency regularization can be easily applied to almost any type of network architecture. 
Furthermore, as features related to computer vision tasks are closely correlated to images,
many parameters (from 90\% to 99.99\%) in the frequency domain can be set to 0 without an obvious decrease in network performance. 
%In the most extreme condition, over 99.99\% of parameters in a UNet can be truncated with less than 2\% of dice score decrease.
%We conducted comprehensive experiments on various state-of-the-art network architectures to demonstrate this.
The proposed frequency regularization has a few desirable properties:
\begin{itemize}
	\item Generality: The proposed frequency regularization can be applied to almost any type of network architecture, as it is designed for tensors. Given this, we are able to evaluate the proposed frequency regularization on various state-of-the-art network architectures including LeNet, AlexNet, VGG, ResNet, ViT, UNet, GAN and VAE.
	\item Effectiveness: A large number of parameters can be truncated in the networks trained with the proposed frequency regularization, since it is widely recognized that high frequency information is unimportant for image data and the features learned by networks are closely correlated to images.
%	\item Convenience: The proposed frequency regularization can be easily applied to any learnable parameter of a network because the inverse discrete transformation does not change the size of tensors representing network parameters. 
	\item Lossless property: There is no need to worry that the proposed frequency regularization will decrease network performance, since the inverse discrete cosine transform (IDCT) is usually considered as a lossless transformation. When no parameter is truncated, the proposed frequency regularization has almost no effect on the network performance.
\end{itemize}

\section{related work}
\label{sec_rel}
Frequency information has been widely used in convolutional neural networks for improving performance \cite{chen2020frequency, ciurana2019hybrid, luo2022frequency, rhee2022lc, saldanha2021frequency, shi2021transform, ulicny2019harmonic, wangVTCLFCVisionTransformer2022, zhang2021compconv, 2022_ACMMM_3548314, 2022_ACMMM_3548303,2022_ACMMM_3548267, 2022_ACMMM_3547924} pruning and network compression 
\cite{chenDiscreteCosineTransform2023,goldbergRethinkingFunFrequencydomain2020,	dziedzicBandlimitedTrainingInference2019,shaoMemoryefficientCNNAccelerator2021,	wangFiltersCompactFeature2017,2020_cvpr_9157408,NIPS2016_36366388,2019_TPAMI_8413170,NIPS2017_8a1d6947,hou2022chex,lin2020hrank,li2016pruning,gao2022disentangled,fang2023depgraph}, or increasing the detection accuracy \cite{jiaExploringFrequencyAdversarial2022, zhongDetectingCamouflagedObject2022, shuCrowdCountingFrequency2022}.
For instance, 
Wang et al. \cite{2022_ACMMM_3548267} represented object edge and smooth structures using high and low frequency information, respectively. 
Mi et al. \cite{2022_ACMMM_3548303} split channel recognition networks into frequency domains, 
while Rippel et al. \cite{rippelSpectralRepresentationsConvolutional2015} proposed a fully spectral representation of networks parameters.
Buckler et al. \cite{bucklerDensePruningPointwise2021} introduced a similar idea with learnable parameters to determine the importance of different channels. 
Besides this, Han et al. \cite{han2016deep} compressed the networks with pruning, quantization and Huffman coding to achieve excellent results for network compression.
Similarly, Wang et al. \cite{wangCnnpackPackingConvolutional2016, 2019_TPAMI_8413170} combined DCT, k-means, quantization and Huffman coding  to compress the network parameters.
%However, this method does not perform well on small convolutional kernels.
Chen et al. \cite{chenCompressingConvolutionalNeural2016} utilized DCT and hashing to assign high frequency to less hash buckets to achieve compression.
Wang et al. \cite{2021_ICLR_dosovitskiy2021an} compress the vision transformer by removing the low-frequency components.
Additionally, several works have proposed using the frequency representation of convolution kernels to prune less important channels \cite{linEZCropEnergyZonedChannels2022, zhangFilterPruningUniqueness2023, liuFrequencydomainDynamicPruning2018}. 
Most of these previous works have primarily focused on the frequency domain representation of the input image, feature maps, or network parameters,
but ignored the learning ability of the network parameters during training process.
Since the network performance is usually sensitive to changes in a few key parameters, it cannot be guaranteed that these key parameters are located in the low frequency domain. 
As a result, these methods have only dropped 30\% to 95\% of parameters with minimal accuracy loss.
In particular, although a few methods \cite{2019_TPAMI_8413170, han2016deep} claimed around 50$\times$ of compression rate, but such rate is improved by quantization and entropy coding. 
The actual proportion of dropped parameters are still ranged from 30\% to 95\%.
In contrast, the proposed frequency regularization approach restricts network parameters to the frequency domain, and high frequency elements are dynamically truncated during network training. 
This allows us to have much higher dropping rates of 90\% to 99.99\%.

The proposed frequency regularization approach is somewhat related to the DCT-Conv method proposed in \cite{2020_IJCNN_9207103}, as both approaches maintain parameters in the frequency domain and utilized IDCT. 
However, there are several key differences between the two methods. 
First, the proposed frequency regularization can be applied to any layer with learnable parameters, including but not limited to convolution layers, fully connected layers, and transposed convolution layers.
Therefore, we are able to evaluate the proposed frequency regularization on various of network architectures including Alexnet, VGG, Resnet, ViT, UNet, GAN and VAE.
In contrast, DCT-Conv is a convolution layer with DCT, which are only evaluated on VGG and Resnet. 
Second, DCT-Conv randomly drops parameters in the frequency domain, which is not ideal since the components in low frequency and high frequency are considered as equally important.
The ability to drop parameters of DCT-Conv is thus limited. Similar limitation is also shown in BA-FDNP method \cite{liuFrequencydomainDynamicPruning2018} in which the coefficients in frequency domain are used for pruning.
Instead, in our approach, we truncate the tail parts of the parameters, as the high frequency components have been shown to be non-critical in features related to images.
It is the most important difference between the proposed approach and these previous methods \cite{liuFrequencydomainDynamicPruning2018, 2020_IJCNN_9207103, 2019_TPAMI_8413170, linEZCropEnergyZonedChannels2022, zhangFilterPruningUniqueness2023} , and it is also the key that the proposed approach can achieve promising compression rate on various of network architectures.
%including LeNet, Alexnet, VGG, Resnet, ViT, UNet, GAN, and VAE,

\section{Frequency Regularization and Tail-Truncation}
When convolutional neural networks are applied to computer vision applications, the learned features are closely related to the training images. 
In fact, as shown by the visualized filters in AlexNet \cite{alexnet}, 
the learned features actually look like real images. 
Based on this insight, it is reasonable to assume that the low frequency components of network parameters is more important than the high frequency components. 
Unfortunately, because of the poor interpretability of convolutional neural networks, 
a network's performance may be highly sensitive to changes of the values in a few key parameters, which are not always located in the low frequency domain. 
This limits the proportion of dropped parameters in pre-trained models. 
Thus, we focus on restricting the number of non-zero parameters during the training process, 
and frequency regularization is thus proposed. 
It should be noted that network regularizations are typically introduced to prevent overfitting in neural networks \cite{goodfellow2016deep}. 
In this paper, however, the definition of regularization is extended to include modifications or restrictions applied to networks with a particular purpose, such as restricting the number of non-zero parameters.

The idea behind the proposed frequency regularization is quite straightforward. 
Instead of maintaining the tensors of network parameters in the spatial domain, they are maintained in the frequency domain. 
This allows the tail elements of tensors to be truncated by a dot product with a zigzag mask matrix. 
During network training, the frequency tensors are input into the inverse discrete cosine transform (IDCT) to reconstruct the spatial tensors.
Then, these spatial tensors are used as the regular learning kernels of convolutional neural networks for learning features. 
Since the IDCT process is differentiable, the parameters maintained in the frequency domain can be correctly updated during backpropagation.
Moreover, as the IDCT can be used for tensors with any dimension without changing their size, 
the proposed frequency regularization can be implemented for any layer involving tensor operations, including convolution layers, fully connected layers, and transposed convolution layers. 
We first introduce the proposed frequency regularization for 1D tensors, and then discuss the implementation of regularization in higher dimensions.

Assume the 1D tensors of learning kernels to be $T(x) \in \mathbb{R}^{1 \times D}$ where $x$ is the index. $T(x)$ represents the learnable parameters in the frequency domain which have been updated during training.
$T(x)$ is first computed with a zigzag mask $\mathds{1}_{x < \epsilon}(x)$ for tail-truncation, then input into the IDCT to reconstruct the spatial tensor $W(y) \in \mathbb{R}^{1 \times D}$ which can be the regular learning kernel in a 1D convolution layer.
Mathematically:
\begin{equation}
	\begin{aligned}
		W(y) & = \mathcal{F}_{\epsilon}(T(x)) = \text{IDCT} \left( \bigcup_{x \in G_x}  T(x) \cdot \mathds{1}_{x < \epsilon}(x)  \right)  \\
			& = \sum_{x \in G_x} T(x) \cdot \mathds{1}_{x < \epsilon}(x) \cdot \text{cos}\left[ \frac{\pi}{D} \left( y + \frac{1}{2} \right ) x \right] \\
			& = \sum_{x < \epsilon} T(x) \cdot  \text{cos}\left[ \frac{\pi}{D} \left( y + \frac{1}{2} \right ) x \right],
	\end{aligned}
\end{equation}
where $x,y \in [0 \ D\!\!-\!\!1] \cap \mathbb{N} = G_x$ are the indices of tensor $T(x)$ and $W(y)$ respectively. 
$T(X)$ is the actual tensor in the frequency domain and $W(y)$ is the corresponding reconstructed tensor in the spatial domain.
$\text{IDCT}(\cdot)$ is the inverse discrete cosine transform.
During the implementation, DCT-III which is the inverse of the widely used DCT-II is used.
$\epsilon$ is the threshold value to control the truncation ratio.
$\mathds{1}_{x < \epsilon}(x)$ is a binary mask to make the truncation process differentiable.
Note that we removed the $\frac{T(0)}{2}$ in the above DCT-III since the learnable parameters can adaptively adjust this constant component.

Since the high dimensional IDCT can be decomposed into several 1D IDCTs,
so high dimensional frequency regularization can also be expressed by several 1D frequency regularizations.
Assume the two N-dimensional tensors in the frequency domain and spatial domain to be $T(\vec{x}), W(\vec{y}) \in \mathbb{R}^{D_1 \times D_2, \cdots, D_N } $ where $\vec{x} = \{x_1, x_2, \cdots, x_N \} $ and $\vec{y} = \{y_1, y_2, \cdots, y_N \} $ are index vectors, $ x_i, y_i \in [0, D_i\!\!-\!\!1] \cap \mathbb{N}$.
Then high dimensional frequency regularization $\mathcal{F}^{N}_{\epsilon}$ is:
\begin{equation}
	\begin{aligned}
		W(\vec{y}) & = \mathcal{F}_{\epsilon}^{N}(T(\vec{x})) = \text{IDCT}^{N} \left( \bigcup_{ \vec{x} \in G_{\vec{x}} }  T(\vec{x}) \cdot \mathds{1}_{|\vec{x}|_1 < \epsilon}(\vec{x})  \right)  \\
			& = \sum_{\vec{x} \in G_{\vec{x}} } T(\vec{x}) \cdot \mathds{1}_{|\vec{x}|_1 < \epsilon}(\vec{x})  \prod_{i = 1}^{N}   \text{cos}\left[ \frac{\pi}{D_i} \left( y_i + \frac{1}{2} \right ) x_i \right] \\
			& = \sum_{|\vec{x}|_1 < \epsilon} T(\vec{x})   \prod_{i = 1}^{N}   \text{cos}\left[ \frac{\pi}{D_i} \left( y_i + \frac{1}{2} \right ) x_i \right]
	\end{aligned}
\end{equation}
where $\text{IDCT}^N$ is the N-dimensional inverse discrete cosine transform, which can be easily implemented by N IDCTs in different dimensions.
$\epsilon$ is the threshold value to control the truncation ratio. $|\vec{x}|_1 =  \sum_{x = 1}^{N}|x_i|$ is the $L_1$ norm of the index vector $|\vec{x}|$.
The indicator function $\mathds{1}_{|\vec{x}|_1 < \epsilon}(\vec{x})$ is used to approximate the zigzag binary mask for truncating parameters.

Frequency regularization is proposed for tensors.
Thus, it can be applied to any layer of a convolutional neural network, such as the convolution layer or fully connected layer.
After applying the proposed frequency regularization,
the formula of the convolution process can be re-expressed as:
\begin{equation}
	Z = WX + B \Rightarrow Z = \mathcal{F}_{\epsilon}^N(T_W)X + \mathcal{F}_{\epsilon}(T_B),
\end{equation}
where $T_W$ and $T_B$ are the tensors representing the weight and bias of the convolution layer.
Usually, the size of the tensor in the convolution layer is 4D representing the number of kernels, input channel and kernel size.
Based on the requirements of different applications, the proposed frequency can be 1D, 2D, 3D or 4D which can be controlled by users.
In particular, the 4D frequency regularization gives us the highest compress rate, which has been used in the proposed evaluation experiments.

\begin{table*}[ht!]
	\caption{Comparison between the proposed approach and DeepCompress\cite{han2016deep} on the MINIST dataset \cite{lecun1998gradient}.}
	\vspace{-10pt}
	\setlength{\tabcolsep}{0.27em}
	\label{tab_compare}
		\begin{tabular}{l|rrr|rrr|rrr}
			\toprule
			\makecell{} &  \multicolumn{3}{c|}{ DeepCompress\cite{han2016deep}}  & \multicolumn{3}{c|}{ DynSurgery \cite{guo2016dynamic} and BA-FDNP \cite{liuFrequencydomainDynamicPruning2018}	}	 & \multicolumn{3}{c}{ The Proposed Approach}							 \\ 
			\makecell{} & \makecell{ Top-1 \\ Accuracy} & \makecell{Compress Rate} & \makecell{ Number of \\ Parameters}   & \makecell{ Top-1 \\ Accuracy } & \makecell{ Compress Rate}	& \makecell{ Number \\ of Parameters}	 & \makecell{ Top-1 \\ Accuracy } & \makecell{ Compress Rate}	& \makecell{ Number \\ of Parameters}						 \\ 
			\midrule\midrule
			LeNet300-ref		& 98.36\% 		&  100\%(1$\times$)		&	266K	&  97.72\% 	& 100\%(1$\times$)		&  266K 	&  98.15\% 	& 100\%(1$\times$)		&  266K 		\\
			LeNet300-v1			& 98.42\%		&  2.5\%(40$\times$)	&	21k(float6)		&  98.01 \% 	& 1.8\%(56$\times$)		& 4.8k				&  97.11\% 	& 0.88\%(114$\times$)	& 4238(float16)			\\	
			LeNet300-v2			&				&						&					&				&						&					&  95.69\% 	& 0.53\%(188$\times$)	& 2816(float16)			\\
			\cdashline{1-10}                                                                                                                                                                                                                                                     
			LeNet5-ref			& 99.20\%		&	100\%(1$\times$)	&	431K	&  99.09\% 	& 100\%(1$\times$)		&	431K	&  98.78\% 	& 100\%(1$\times$)		&	431K		\\
			LeNet5-v1			& 99.26\%		&	2.56\%(39$\times$)	&	34K(float6)		&  99.09\% 	& 0.92\%(108$\times$)	& 	4.0k			&  98.35\% 	& 0.89\%(112$\times$)	& 	7668(float16) 		\\
			\cline{1-10}
			\makecell{} &  \multicolumn{3}{c|}{ }  & \multicolumn{3}{c|}{  BA-FDNP \cite{liuFrequencydomainDynamicPruning2018}	}	 & \multicolumn{3}{c}{ The proposed approach }							 \\ 
			\cdashline{5-10}
			LeNet5-v2			&				&						&					& 99.08\%				& 	0.67\%(150$\times$)		& 2.8K 	&  97.52\% 	& 0.09\%(1110$\times$)	&   776(float16)		\\
			\bottomrule
		\end{tabular}%
		\vspace{-10pt}
\end{table*}
\subsection{Dynamic tail-truncation:} 
According to the evaluation experiments we proposed in \refsec{sec_exp}, usually over 99\% of parameters in convolutional neural networks can be truncated without an obvious decrease in accuracy. 
However, there is also a serious issue that the remaining parameters may not have suitable gradients for backpropagation. 
In this condition, the training loss of the network restricted by the proposed frequency regularization sometimes may not change for hundreds of training epochs.
This issue has occurred many times in our evaluation experiments. 
To address this problem, we propose the dynamic tail-truncation strategy. 
Instead of directly setting over 99\% of the parameters to 0, 
this strategy continuously sets a few tail elements to 0 in every training epoch.
In particular, as the total number of truncated parameters increases, the number of parameters truncated in each training epoch decreases.
Mathematically, the ratio of truncated parameters is controlled by following function:
\begin{displaymath}
	\beta_n = \beta_{n - 1} - \gamma (\beta_{n-1} - \epsilon)
\end{displaymath}
where $n$ is the index of training epoches. $\beta_{n-1}, \beta_{n} \in [0, 1]$ is the ratio of non-zeros parameters in epoch $n-1$ and $n$. $\gamma$ is the user parameter to control the speed of truncating parameters. During our evaluation experiment, $\gamma=0.01$ is used.
$\epsilon$ is the user parameter to control the mininum ratio of non-zero parameters in the network.
By changing the value of $\epsilon$, we can control the percentage of parameters that will be truncated in a network.
For example, $\epsilon=0.01$ means that around $1 - \epsilon = $ 99\% of parameters will be truncated.
Although the dynamic tail-truncation strategy requires extra training epochs, 
it results in a more stable training result when the minimum ratio $\epsilon$ is very small.
This strategy has been used for all the evaluation experiments in this paper.

\section{Experiments}
\label{sec_exp}
In this section, we evaluate the proposed frequency regularization on several classical state-of-the-art network architectures including LeNet, Alexnet, VGG, ResNet, ViT, UNet, GAN and VAE on several standard datasets.
During the evaluation, we first capture the accuracy of the original networks which is used as the reference accuracy to compare with the ones of networks restricted by the proposed frequency regularization.
In particular, the implementation provided by authors or popular Github repositories are used.
Then, these networks are re-trained with the proposed frequency regularization to compare the original networks.
Since the parameters of networks are zigzag truncated in the proposed frequency regularization, we only need to save the location of the boundary between non-zero elements and zero elements as well as the size of tensors.
Thus, the compress rate can be easily computed by dividing the number of non-zero parameters by the total number of parameters.
Note that we did not consider the bias in the convolution layers, fully connected layers and transposed convolution layers during our evaluation,
since the bias only represents a small portion of network parameters and most of the previous researchers also ignored them.
Similarly, the parameters in batchnorm layers are also ignored with for same reason.
Given the page limitation, we only demonstrate the total number of non-zero parameters in networks. 
More detailed are presented in the supplementary material. All the experiments are performed on GTX A4000 with 16 GB of video memory.
The source code as well as our pre-trained model with frequency regularization will be made available following the acceptance of this paper.
\begin{table}[ht!]
	\caption{Evaluation of the proposed frequency regularization on Alexnet \cite{alexnet}, VGG \cite{Simonyan15}, ResNet \cite{he2016deep} and ViT \cite{2021_ICLR_dosovitskiy2021an} using CIFAR10 dataset \cite{krizhevsky2009learning}.}
	\vspace{-10pt}
	\label{tab_imgclass}
		\begin{tabular}{@{ }l@{ }|@{ }r@{ }r@{ }r@{ }}
			\toprule
			%\makecell{} &  \multicolumn{2}{c}{ Deep Comp\cite{han2016deep}}  & \multicolumn{2}{c}{ The proposed approach}									 \\ 
			\makecell{}  & \makecell{ Top-1 Accuracy} & \makecell{ Compress Rate} & \makecell{Number of \\ Parameters}						 \\ 
			\midrule\midrule
			%\midrule
			AlexNet-ref			& \makecell{ 76.45 \% }				&		100\%(1$\times$)		&		57,035,456													\\
			AlexNet-v1			& \makecell{ 77.44 \% }				&	 1\%(100$\times$)		&		570,365									 \\
			AlexNet-v2			& \makecell{ 70.46 \% }				&	 0.1\%(1000$\times$)		&		57,364								 \\
			AlexNet-v3			& \makecell{ 59.22 \% }				&	0.0025\%(40509$\times$)		&	1,408							 \\
			AlexNet-v4			& \makecell{ 58.55 \% }				&	0.00123\%(81018$\times$)		&	1,408(float16)							 \\
			\cdashline{1-4}                                                                                             
			VGG16-ref			& \makecell{ 85.84 \% }				&	100\%(1$\times$)		&	20,024,000												\\
			VGG16-v1			& \makecell{ 84.77 \% }				&	1\%(100$\times$)		&	200,578									 \\
			VGG16-v2			& \makecell{ 73.01 \% }				&	0.1\%(987$\times$)		&	20,302									 \\
			VGG16-v3			& \makecell{ 65.99 \% }				&	0.0102\%(9775$\times$)		&	2,048							 \\
			%			VGG16-v2			& \makecell{ 10.00 \% }				&	 19750$\times$		&	2,048(float16)							 \\
			\cdashline{1-4}                                                                                             
			ResNet18-ref		& \makecell{ 85.43 \% }				&	100\%(1$\times$)		&	11,162,624												\\
			ResNet18-v1			& \makecell{ 85.34 \% }				&	 1\%(100$\times$)		&	111,901											 \\
			ResNet18-v2			& \makecell{ 83.23 \% }				&	 0.1\%(1000$\times$)		&	12,058									 \\
			ResNet18-v3			& \makecell{ 77.64 \% }				&	 0.024\%(4156$\times$)		&	2,688							 \\
%			ResNet-v2			& \makecell{ 10.00 \% }				&	 8312$\times$		&	2,688(float16)							 \\
			\cdashline{1-4}
			ViT-ref			& \makecell{ 78.94 \% }				&	 100\%(1$\times$)		&		9,500,672							 \\
			ViT-v1			& \makecell{ 78.98 \% }				&	 9.77\%(10$\times$)		&		928,478						 \\
			ViT-v2			& \makecell{ 75.05 \% }				&	 0.83\%(120$\times$)		&		157,848(float16)						 \\
			\bottomrule
		\end{tabular}%
		\vspace{-10pt}
\end{table}
\subsection{Comparison with state-of-the-art methods}
To the best of our knowledge,
there is no work which is directly related to the proposed approach on restricting the information redundancy during network training.
The closest work we could find was proposed for network compression by pruning, such as DeepCompress \cite{han2016deep}, DynSurgery \cite{guo2016dynamic} or BA-FDNP \cite{liuFrequencydomainDynamicPruning2018}.
Although there are a few newer methods proposed in \cite{2019_TPAMI_8413170,2019_TPAMI_8416559,2020_IJCNN_9207103, 2017_TPAMI_7546875,fang2023depgraph, hou2022chex,lin2020hrank, liuFrequencydomainDynamicPruning2018},
their advantage on compress rate is not very obvious and almost all of them work on networks pre-trained on large datasets.
which makes the comparisons become very expensive considering computational resources.
Besides, among all these previous approach, BA-FDNP \cite{liuFrequencydomainDynamicPruning2018} claimed the highest compress rate which is 150$\times$.
Thus, we compare the proposed approach with DeepCompress \cite{han2016deep} ,DynSurgery \cite{guo2016dynamic} and BA-FDNP \cite{liuFrequencydomainDynamicPruning2018}.
The comparison results are shown in \reftab{tab_compare}. In particular, the top-1 accuracy \cite{alexnet} is used as evaluation metric.
The proposed approach achieves much higher compress rate without obvious decrease in top-1 accuracy.
For example, in LeNet300-v1, the proposed approach achieves 112$\times$ the compress rate with less than 1\% of cost in top1-accuracy.
Furthermore, it achieves 1110 $\times$ compress rate in LeNet5-v2 with only 2\% of top1-accuracy decrease, which is much higher than the DeepCompress \cite{han2016deep}, DynSurgery \cite{guo2016dynamic}, BA-FDNP \cite{liuFrequencydomainDynamicPruning2018} as well as those following works \cite{2019_TPAMI_8413170,2019_TPAMI_8416559,2020_IJCNN_9207103, 2017_TPAMI_7546875,fang2023depgraph, hou2022chex,lin2020hrank, liuFrequencydomainDynamicPruning2018}.
It should be noted that the BA-FDNP \cite{liuFrequencydomainDynamicPruning2018} proposed to do pruning in frequency domain which is related to the proposed approach.
However, as we mentioned in \refsec{sec_rel}, instead of truncating the tail parts of tensor, BA-FDNP \cite{liuFrequencydomainDynamicPruning2018} utilized the coefficients matrix for pruning which limits their compress rate.
As a result, BA-FDNP only has 150$\times$ compress rate on LeNet5-v2, but the proposed can achieve around 1110 $\times$ reduction with less than 2\% of top-1 accuracy decrease.
It should be noted that BA-FDNP \cite{liuFrequencydomainDynamicPruning2018} utilized data argumentation to improve top-1 accuracy.
It also applied the pre-trained model for initialization and retrain their model for 20k iterations for searching the highest top-1 accuracy model,
However,
%the training accuracy of the proposed approach is 100\%, and we didn't prevent the network from overfitting by using data argumentation technique which is utilized in BA-FDNP \cite{liuFrequencydomainDynamicPruning2018}.
%In addition, BA-FDNP applied the pre-trained model for initialization and retrain their model for 20k iterations for searching the highest top-1 accuracy model,
none of these techniques for accuracy improvement are applied in the proposed approach, with the consideration of our computation resource.
%Unfortunately, 
Both LeNet300 and LeNet5 are quite simple architectures and the MINIST dataset is also not a very challenging dataset.
In order to thoroughly evaluate the proposed approach,
we further apply the proposed frequency regularization on various state-of-the-art network architectures in the remaining sections.

\subsection{Image classification} 
\label{sec_class}
Image classification has been studied for decades in computer vision.
There are many excellent architectures that have been proposed in this field.
Thus, we evaluate the performance of the proposed frequency regularization on  Alexnet \cite{alexnet}, VGG16 \cite{Simonyan15}, ResNet18 \cite{he2016deep} and ViT \cite{2021_ICLR_dosovitskiy2021an}.
During evaluation, the cifar10 dataset \cite{krizhevsky2009learning} is used for training and testing and top1-accuracy is used as the evaluation metric.
Note that we did not apply tuning techniques or pre-trained model during evaluation considering training time.
The accuracy scores can be higher once some training techniques such as dynamic learning rate or data argumentation is utilized, which has been widely used in previous approaches \cite{2019_TPAMI_8413170,2019_TPAMI_8416559,2020_IJCNN_9207103, 2017_TPAMI_7546875,fang2023depgraph, hou2022chex,lin2020hrank, liuFrequencydomainDynamicPruning2018}.

The evaluation results are shown in \reftab{tab_imgclass}.
AlexNet-ref which is the original Alexnet \cite{alexnet} achieves 76.45\% top-1 accuracy.
After 99\% parameters have been truncated, AlexNet-v1 achieves 77.44\% top-1 accuracy which is a little bit higher than the original one.
This improvement comes from the fact that the tail-truncation can be considered as a regularization to prevent a network from overfitting just like the dropout layer.
When only 0.1\% of parameters are kept in the frequency domain, AlexNet-v2 has 70.46\% in top-1 accuracy,
and the compress rate becomes 1000$\times$.
Furthermore,
in the extreme case where only 1408 non-zero parameters are kept in AlexNet, AlexNet-v3 still achieves 59.22\%.
We also applied the network on the half float condition, and AlexNet-v4 consisting of 1408 float16 numbers achieves 58.55\% with the compress rate increasing to 81018$\times$,
which is a very interesting discovery.
Similar results are also observed in VGG16 and ResNet18.
In particular,
ResNet18-v3 with only 2688 parameters achieves 77.64\%, which is even higher than the original AlexNet-ref with 57M parameters.
We also evaluate the proposed frequency regularization on ViT consist of self-attention layers which are actually not very suitable frequency domain transformation.
Previous work \cite{wangVTCLFCVisionTransformer2022, fang2023depgraph} related to transformer only achieves around 50\% of pruning ratio.
For the proposed approach, there is around 75\% top-1 accuracy in ViT-v2 in which over 98\% of parameters are truncated.
But when only 90\% of parameters are truncated, the ViT-v1 has no accuracy decrease, which demonstrate the generality of the proposed approach.
The evaluation of proposed approach on these networks clearly demonstrates that the information redundancy inside networks can be restricted well by the proposed frequency regularization.
However, the resolution of images in CIFAR10 is not very high,
which may raise a concern that the proposed frequency regularization only works well on small images.
Thus, we evaluated the proposed technique on UNet for high resolution image segmentation.
\begin{table}[ht!]
	\caption{Evaluation of the proposed frequency regularization on UNet for image segmentation tasks using Carvana Image Masking Challenge Dataset \cite{iglovikov2018ternausnet}.}
	\vspace{-10pt}
	\setlength{\tabcolsep}{0.27em}
	\label{eva_unet}
%	\resizebox{0.49\textwidth}{!}{%
		\begin{tabular}{l|rrr}
			\toprule
			\makecell{} & \makecell{ Dice Score} & \makecell{Compress Rate}  & \makecell{Number of Parameters}						 \\ 
			\midrule\midrule
			%\midrule
			UNet-ref		& 99.13 \% 		& 100\%(1$\times$)				&		31,043,586 															\\
			UNet-v1			& 99.51 \%		&	 1\%(100$\times$)			&			310,964												 \\
			UNet-v2			& 99.37 \%		&	 0.1\%(1000$\times$)			&		31,096												 \\
			UNet-v3			& 98.86 \%		&	0.0094\%(10573$\times$)			&		2,936									 \\
			\cdashline{1-4}
			UNet-v4			& 97.19 \%		&   0.0012\%(81801$\times$)	        &      759(float16)								\\
			\bottomrule
		\end{tabular}%
		\vspace{-10pt}
\end{table}
\subsection{Image Segmentation}
\label{sec_seg}
\begin{figure}[!t]
	\includegraphics[width=\linewidth]{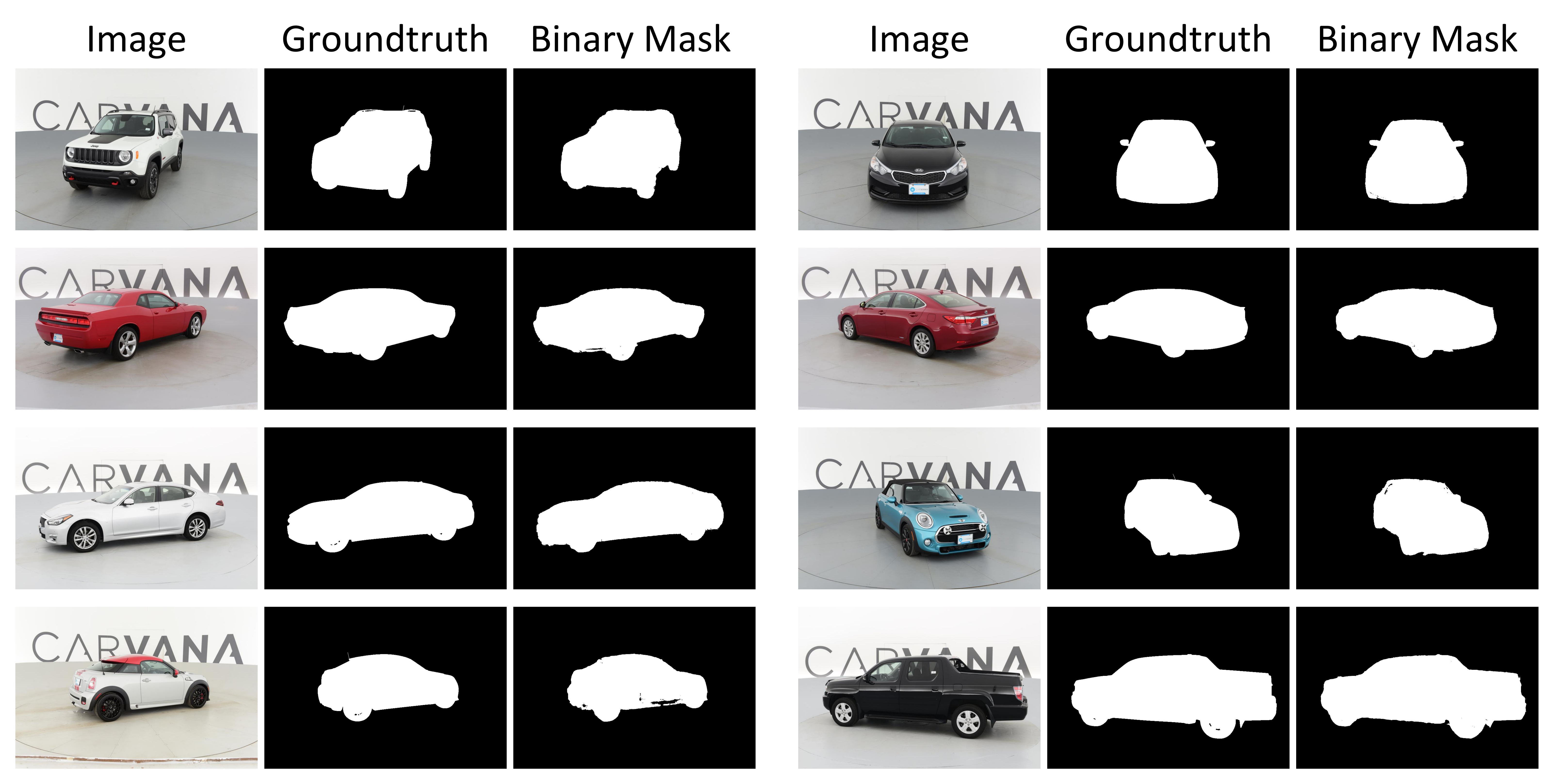}
	\vspace{-20pt}
	\caption{The segmentation results of UNet parameters restricted by the proposed frequency domain, with over 99.99\% of parameters truncated, for the Carvana Image Masking Challenge Dataset \cite{iglovikov2018ternausnet}.}
    \label{fig_UNET}		
\end{figure}
After a comprehensive literature review, we found a limited number of methods focussing on prunning or compressing segmentation networks, even DepGraph \cite{fang2023depgraph}, which is the latest prunning method proposed for any structural prunning, was not evaluated on segmentation networks.
One reason maybe the sense that the information redundancy for segmentation networks should be lower than image classification networks.
However, the proposed frequency regularization is supposed to work well on any features related to images and segmentation networks should be one of them.
We thus evaluate the proposed frequency regularization on UNet using the Carvana Image Masking Challenge Dataset \cite{iglovikov2018ternausnet}, which is a popular dataset for segmentation challenge in Kaggle competitions.
In particular, the Dice Score is used for evaluation.
\begin{table}[ht!]
	\caption{Evaluation of the proposed frequency regularization on generative adversarial network (GAN) and variational autoencoder (VAE).}
	\vspace{-10pt}
	\setlength{\tabcolsep}{0.27em}
	\label{tab_imggen}
%	\resizebox{0.49\textwidth}{!}{%
		\begin{tabular}{l|rrr}
			\toprule
			%\makecell{} &  \multicolumn{2}{c}{ Deep Comp\cite{han2016deep}}  & \multicolumn{2}{c}{ The proposed approach}									 \\ 
			\makecell{} & \makecell{ Fid value} & \makecell{Compress Rate} &  \makecell{Number of Parameters}							\\ 
			\midrule\midrule
			%\midrule
			GAN-ref	& \makecell{244 }				& 100\%(1$\times$)				&		 6,334,464 							\\
			GAN-v1	& \makecell{248 }				& 1.02\%(98 $\times$)			&		64,383								\\
			\cdashline{1-4}                                                                                             
			VAE-ref	& \makecell{81 }				& 100\%(1$\times$)				&		971,200								\\
			VAE-v1	& \makecell{94 }				& 10\%(10$\times$)				&		97,134								\\
			\bottomrule
		\end{tabular}%
		\vspace{-10pt}
\end{table}
Actually, compared to image classification networks, UNet architecture usually consists of pure convolution layers with on bias which are more suitable for the proposed frequency regularization.
As shown in \reftab{eva_unet}, the orignal UNet containing 31M parameters achieves 99.31\% Dice score.
Once over 99\% parameters have been truncated in the frequency domain, UNet-v1 achieves 99.51 \% Dice score.
Moreover, UNet-v3  achieves promising results with 98.86\% in Dice score, in which only 2936 parameters are kept.
Finally, we tested the proposed approach on the most extreme condition in which only 759 float16 parameters are kept in UNet-v4.
In particular, we also disabled all the bias in the networks and learning parameters in batchnorm, which guaranteed the UNet-v4 only has 759 non-zero float16 parameters.
Surprisingly, the UNet-v4 still achieves 97.19\% in Dice score.
This is an unbelievable result, a UNet with only 759 parameters can achieve around 97\% in Dice score for the Carvana Image Masking Challenge Dataset.
Note that there are over 5000 images with 959 $\times$ 460 resolution which is much higher than the resolutions of images in CIFAR10 \cite{krizhevsky2009learning}.
We double checked the non-zero parameters in every layer in UNet to make sure that this conclusion is correct, and the visualized segmentation mask is shown in \reffig{fig_UNET}.
We also include this pre-train UNet into the supplementary materials as well as a few testing images. 
\emph{The original size of the UNet model exceeds 366MB, but our frequency regularization technique reduces it to 40kb.
Additionally, using an entropy compression tool on Ubuntu, we further reduce the size to 4.5kb.}
\begin{figure}[!t]
	\begin{subfigure}[b]{0.45\linewidth}
		\centering
		GAN-ref: 244 Fid value
		\includegraphics[width=0.9\linewidth]{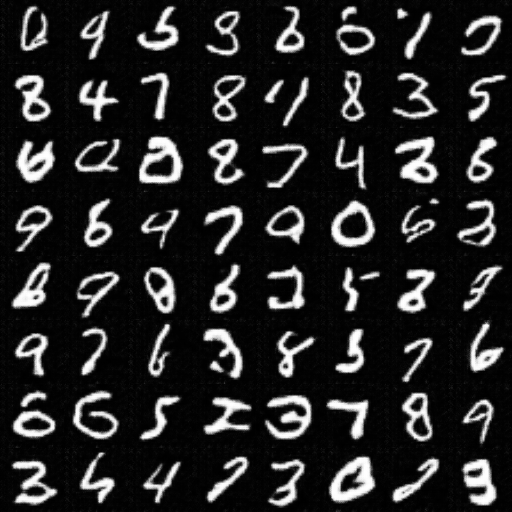}
	\end{subfigure}
	\hspace{4pt}
	\begin{subfigure}[b]{0.45\linewidth}
		\centering
		  GAN-v1: 248 Fid value
		\includegraphics[width=0.9\linewidth]{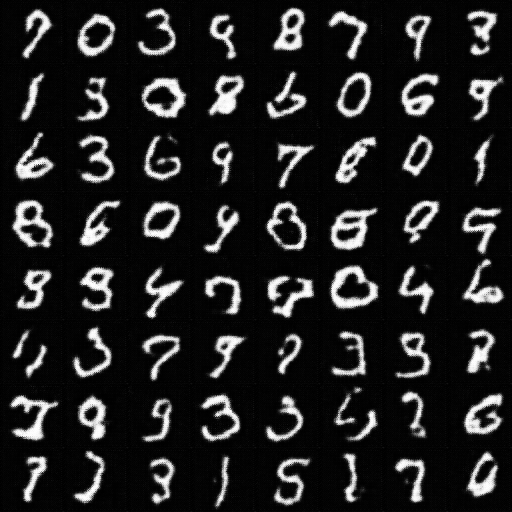}
	\end{subfigure}
	\vspace{-10pt}
	\caption{Comparisons between images generated by the orignal GAN network and GAN with our frequency regularization in which over 99\% of parameters are truncated.}
    \label{fig_gan}		
	\vspace{-10pt}
\end{figure}
\subsection{Image Generation}
\label{sec_gen}
\begin{figure}[!t]
	\begin{subfigure}[b]{0.45\linewidth}
		\centering 
		VAE-ref: 81 Fid value
		\includegraphics[width=0.9\linewidth]{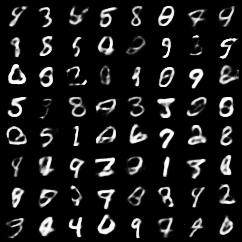}
	\end{subfigure}
	\hspace{4pt}
	\begin{subfigure}[b]{0.45\linewidth}
		\centering
		VAE-v1: 94 Fid value
		\includegraphics[width=0.9\linewidth]{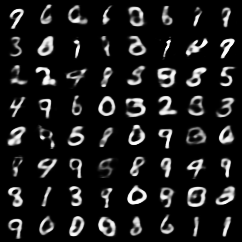}
	\end{subfigure}
	\vspace{-10pt}
	\caption{Comparisons between images generated by the orignal VAE network and VAE with our frequency regularization in which over 90\% of the parameters are truncated.}
    \label{fig_VAE}		
	\vspace{-14pt}
\end{figure}

Compare to classification or segmentation networks, image generation networks such as the generative adversarial network (GAN) or variational autoencoder (VAE) are not very suitable for the proposed frequency regularization, 
since what has been learned by these networks are claimed as distribution information.
However, these distributions are still related to images. Thus, we also evaluate the proposed approach on generative adversarial network and varitional autoencoder.
Since both networks usually require long time for training, we used the MINIST dataset for evaluation and the Fid value \cite{NIPS2016_36366388} is used as the evaluation metric.
%We first apply the frequency regularization on GAN.
As shown in \reftab{tab_imggen}, 
the proposed approach achieves similar results when 99\% of the parameters are truncated compared to the orignal GAN network.
The generated images are also demonstrated in \reffig{fig_gan},
where the images generated by GAN with the proposed frequency are similar to the original GAN.
We also evaluate the proposed approach regularization on VAE.
Since the VAE we used is the simplest version in which there are only two fully connected layers in their encoder and three fully connected layers in their decoder, we only truncate 90\% of the parameters for VAE.
The visual results are shown in \reffig{fig_VAE}.
During the training of VAE and GAN, with or without the proposed frequency regularization, we stop the training once the visual results look good. Actually, with more training epoches, the quality of the generated images can be better.
\subsection{Limitation and Future Work}
The proposed frequency regularization is based on the assumption that the high frequency component is unimportant. This may not work well on the tensors of parameters which are not related to images.
This disadvatange can be seen for the comparison between evaluations on image segmentation in \refsec{sec_seg} and image generation in \refsec{sec_gen},  where the parameters in the generation networks cannot be truncated too much.
In addition, although the parameter of the proposed approach is in the frequency domain, it still needs to be converted into the spatial domain for convolution operations.
This leads to an extra memory cost during networks training.
However, once the networks have been well-trained,
a high compress rate can be achieved which is very useful for network transmission on the Internet.
The memory cost does not exist during testing, since the networks need to be unpacked only once.
Since around 99\% of the parameters in networks restricted by the proposed frequency regularization can be truncated, it is very much possible to speed-up the network inference. This will be the next direction of our research.
\section{Conclusion}
In this paper we proposed frequency regularization to restrict the information redundancy of convolutional neural networks devised for computer vision tasks.
The proposed regularization maintains the tensors of parameters in the frequency domain where the high frequency component can be truncated.
During training, the tail part of tensors are truncated first before being input into the inverse discrete cosine transform to reconstruct the spatial tensors that are used for tensor operations.
In particular, the dynamic tail-truncation strategy was proposed to improve the stability of network training.
We applied the proposed frequency regularization in various state-of-the-art network architectures for evaluation.
Comprehensive experiments demonstrated that between 90\% to 99.99\% of the parameters in the frequency domain can be truncated.
This demonstrates the promising ability of the proposed frequency regularization to restrict the information redundancy in convolutional neural networks for computer vision tasks.

\bibliographystyle{ACM-Reference-Format}
%\bibliography{sample-base}
\bibliography{ref.bib}

\appendix

\end{document}